\documentclass[journal,twoside]{IEEEtran}
\usepackage{amsfonts,bm,booktabs,cite,graphicx,hyperref,multirow,textcomp,url}
\usepackage{algorithm}
\usepackage{subfigure}
\usepackage{algpseudocode}
\usepackage{amsmath}
\usepackage{mathrsfs}
\usepackage{float}
\usepackage{color}
\usepackage{booktabs}
\usepackage{bbding}
\usepackage{adjustbox}
\usepackage[flushleft]{threeparttable}

\makeatletter 
\renewcommand{\@thesubfigure}{\hskip\subfiglabelskip}
 \makeatother

\usepackage{CJKutf8}

\begin{document}
\title{Video Codec-Guided Training-Free Token Compression for Video Large Language Models}
\author{Xihua Sheng,~\IEEEmembership{Member,~IEEE}, 
Ruijie Liu, 
Chang Wen Chen,~\IEEEmembership{Life Fellow,~IEEE}\\
\thanks{Date of current version \today.}
\thanks{X. Sheng and C. Chen are with the Department of Computing, The Hong Kong Polytechnic University, Hong Kong SAR, China (e-mail: xihua.sheng@polyu.edu.hk, changwen.chen@polyu.edu.hk).\par
R. Liu is with the Department of Informatics, Imaging and Data Sciences, The University of Manchester, Manchester, U.K. (e-mail: liuruijie744@gmail.com).
}
\thanks{Xihua Sheng and Ruijie Liu contributed equally to this work.}
\thanks{Corresponding Author: Chang Wen Chen.}
}

\markboth{Journal of \LaTeX\ Class Files,~Vol.~14, No.~8, August~2021}%
{Shell \MakeLowercase{\textit{et al.}}: A Sample Article Using IEEEtran.cls for IEEE Journals}

\maketitle
\begin{abstract}
Video large language models (VLLMs) rely on long visual token sequences to understand complex temporal content, which makes visual token compression important for efficient inference. Existing training-free compression methods mainly reduce the number of tokens sent to the language model, but the compression process itself can still require many visual feature comparisons, especially for intra-frame token protection and inter-frame temporal token merging. This paper proposes a video codec-guided training-free token compression method for VLLMs that uses coding information already available in compressed video as a low-cost search prior. The method aligns reference-frame dependencies, motion vectors, and residuals with the visual token grid. Residual information forms compact candidate pools for token protection, while motion information is propagated
along reference-frame chains to construct localized candidates for temporal token merging between sparsely sampled frames. Loacl visual feature similarity comparison is then used within these candidates to make the final compression decisions,
without modifying the base VLLM. Experiments on three representative VLLMs and four video-understanding benchmarks show that the proposed method achieves
performance comparable to the state-of-the-art training-free baseline FlashVID, and in some settings surpasses it, while substantially reducing the feature comparisons required by token protection and temporal matching. These results demonstrate that codec information can effectively improve the accuracy-complexity tradeoff of video token compression.
\end{abstract}


\section{Introduction}
\label{sec:introduction}
Video large language models (VLLMs) have shown strong perception and reasoning abilities on video understanding tasks~\cite{lin2024videollava,li2024llavaonevision,zhang2024llavavideo,shu2025videoxl,wang2025dynamicvlm}, and have become an important direction in multimodal understanding. They typically encode video frames into visual tokens and feed them, together with text tokens, into a large language model. To handle long videos, complex events, and fine-grained temporal relations, the model needs to process longer visual contexts. However, longer contexts introduce more visual tokens, which substantially increase attention computation, prefilling latency, and KV-cache storage. Therefore, visual token compression is important for improving the inference efficiency of video large language models while preserving their understanding performance.
\begin{figure}[t]
    \centering
    \includegraphics[width=\linewidth]{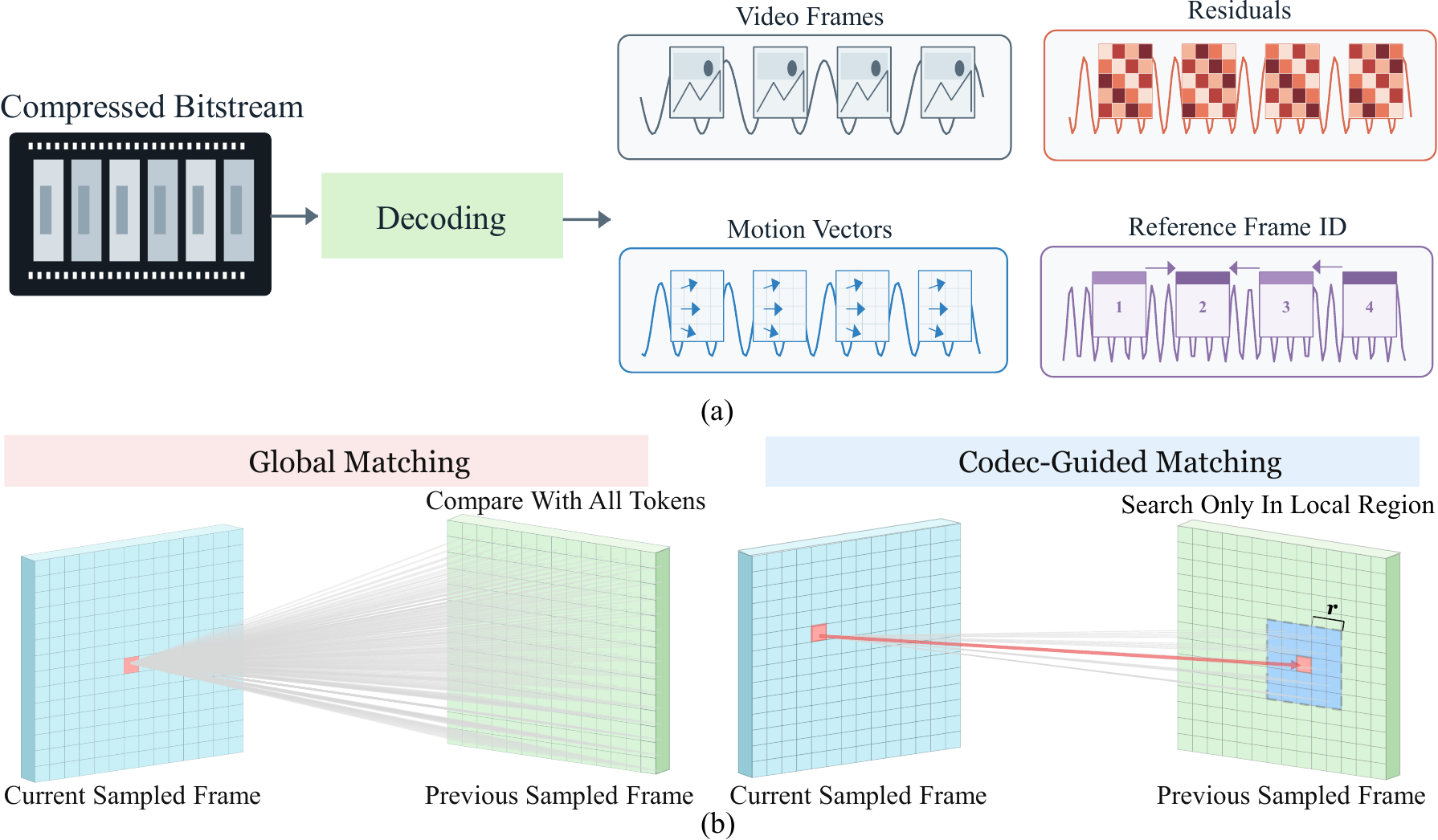}
    \caption{Illustration of video codec information extraction and codec-guided local token matching. (a) Video codec information extraction. (b) Global visual token matching and codec-guided local token matching. }
    \label{fig:global_local_matching}
\end{figure}
Existing VLLM visual token compression methods~\cite{rao2021dynamicvit,liang2022evit,bolya2023tome,chen2024fastv,yang2025visionzip,huang2025prunevid,shen2025fastvid,fan2026flashvid} reduce the number of tokens entering the language model by pruning or merging tokens. They mainly target the length of the compressed token sequence, but the compression decisions themselves also have a computational cost. This issue is particularly important for video. Intra-frame selection must decide which tokens should be retained, while temporal merging must decide whether a token in the current sampled frame has already been represented by other sampled frames. If these decisions rely on global comparisons among visual features, the compression module can introduce many similarity computations. In other words, the efficiency of video token compression depends not only on the final token retention ratio, but also on the number of visual token comparisons required to make the compression decisions.

The cost of these visual feature comparisons is closely related to the structure of video redundancy. Intra-frame redundancy often comes from repeated textures, background regions, or locally similar content, whereas inter-frame redundancy comes from the repeated appearance of the same object or region over time. These two types of redundancy require different token compression decisions: intra-frame token selection needs to identify distinctive or informative tokens, while temporal token merging needs to locate where the same content moves across sampled frames. Fixed-position matching may miss moving objects, whereas global cross-frame search introduces a quadratic number of comparisons. This motivates the use of a low-cost prior to construct a smaller and more reliable candidate region before computing visual feature similarity.

Our key observation is that compressed video already contains rich prior
information about motion and local changes. As illustrated in
Fig.~\ref{fig:global_local_matching}, to remove pixel redundancy, a video codec selects reference frames for coding blocks, estimates motion vectors, and encodes residuals after motion compensation. The reference relation indicates which frame a block depends on, the motion vector indicates where it is predicted from in the reference frame, and the residual reflects the local prediction error that remains unexplained by this prediction. These signals are not designed for downstream semantic tasks, but they describe how redundancy is removed during video coding~\cite{chien2021vvc,wu2018coviar,shou2019dmcnet,zheng2024covifocus,li2022compressedgebd,sheng2024spatial,chen2021cdvsr}. Therefore, they can serve as a low-cost search prior that helps visual token compression avoid unnecessary global comparisons.

However, these coding signals cannot be used directly. Coding blocks and visual tokens have different spatial granularities. A single token may cover multiple coding blocks, and these blocks may have different motion directions and reference frames. A region with a large residual is also not necessarily semantically important, since texture, compression noise, or coding errors may produce similar responses. Thus, coding information is more suitable for generating candidate sets than for directly deciding which tokens should be kept or merged. The final compression decision should still rely on visual feature similarity comparison. Another important challenge comes from sparse frame sampling in VLLMs. A VLLM visually encodes only a small number of sampled frames~\cite{lin2024videollava,li2024llavaonevision,zhang2024llavavideo,shu2025videoxl}, while coding references are defined over the complete video~\cite{bross2021overview,sullivan2012overview}. The frame referenced by a current sampled frame is often not the preceding sampled frame, but an intermediate frame between them. Therefore, a single-step motion vector cannot directly describe token correspondence between two sampled frames. Ignoring intermediate frames breaks the actual reference path, while globally matching all tokens between sampled frames loses the efficiency benefit. To address this issue, motion information should be propagated along the reference-frame dependencies in the complete video, so that multiple short coding dependencies can be combined into long-range candidate positions between sampled frames.

Based on these observations, we propose a video codec-guided training-free token compression method for video large language models. The method does not modify the base VLLM, nor does it retrain the vision encoder or the language
model. It extracts reference-frame dependencies, motion vectors, and residuals from the complete compressed video, and maps these block-level signals to the visual token grid. Based on the aligned coding information, residual-guided token protection first forms a compact candidate pool for visual token selection, while motion-guided temporal merging uses reference-chain propagation to construct localized cross-frame candidates.

The central principle is to use coding information for candidate generation and local visual feature comparisons for verification. Coding signals are cheap, dense, and
available from the video bitstreams, but they are noisy and not task-aware. Visual feature comparisons provides stronger semantic evidence, but global comparison among visual tokens is expensive. By assigning different roles to these two sources of information, as shown in Fig.~\ref{fig:global_local_matching}, our method converts costly global token matching into codec-guided local matching, reducing comparison cost while preserving compression quality.

Our contributions are as follows:
\begin{itemize}
    \item We propose a video codec-guided training-free token compression framework for video large language models. The framework exploits reference-frame dependencies, motion vectors, and residuals already available in compressed videos to enable low-cost visual token compression without modifying the base model.

    \item We develop a residual-guided token protection strategy that uses residual information to form a compact candidate pool, reducing the comparison cost of token protection while preserving informative visual evidence.

    \item We design a motion-guided temporal merging method based on reference-chain propagation. By following reference-frame dependencies and motion vectors through intermediate coded frames, it connects sparsely sampled frames and produces localized candidates for cross-frame token merging, thereby reducing temporal visual-similarity comparisons.
\end{itemize}

\begin{figure}[t]
    \centering
    \includegraphics[width=\linewidth]{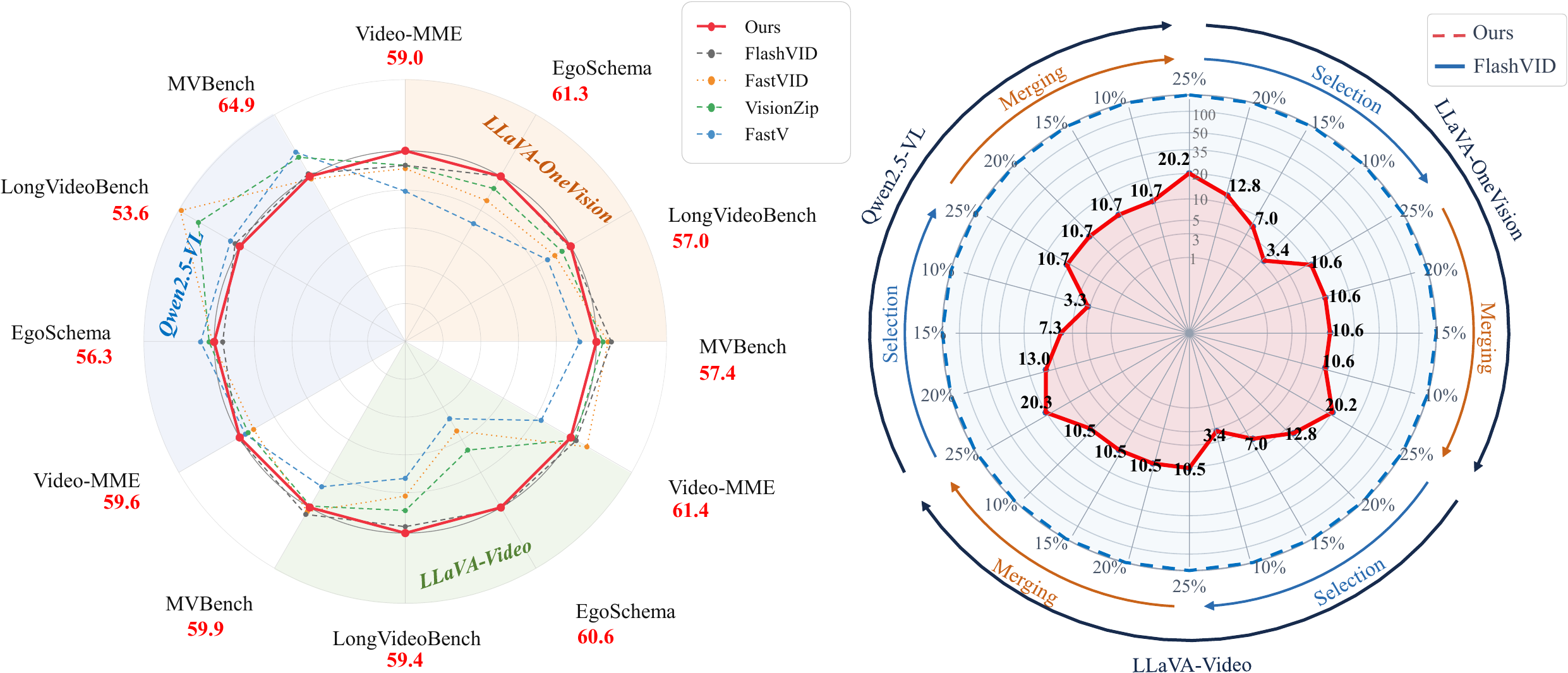}
    \caption{Comparison of the performance and feature-comparison costs of our method and the state-of-the-art baseline FlashVID~\cite{fan2026flashvid}.}
    \label{fig:radar}
\end{figure}

We evaluate the proposed method on three representative video large language models and four video understanding benchmarks. As shown in Fig.~\ref{fig:radar}, our method achieves performance comparable to, and in some cases better than, the state-of-the-art baseline FlashVID~\cite{fan2026flashvid}, while substantially reducing the feature comparisons required for token protection and cross-frame temporal merging. These results show that codec information already present in compressed videos can serve as an effective search prior for improving the accuracy-complexity tradeoff of video token compression without changing the base model.

\section{Related Work}
\subsection{Visual Token Compression}
Visual token compression reduces the length of visual token sequences while preserving task-relevant information. Early work on efficient vision transformers shows that not all visual tokens need to be processed with the same cost. DynamicViT~\cite{rao2021dynamicvit} dynamically prunes less informative tokens during inference, and EViT~\cite{liang2022evit} reorganizes inattentive tokens to accelerate vision transformers. For large vision-language models, FastV~\cite{chen2024fastv} observes that visual redundancy becomes evident after shallow language-model layers and prunes visual tokens according to early-layer attention responses. IVTP~\cite{huang2024ivtp} further incorporates the input instruction into token-importance estimation, while SparseVLM~\cite{zhang2025sparsevlm} uses text-guided sparsification with adaptive layer-wise ratios and token recycling. These pruning-based methods mainly decide which tokens should remain active, using importance signals derived from attention, instructions, or text-visual interactions.

Another line of work compresses visual tokens by merging or aggregating redundant representations. ToMe~\cite{bolya2023tome} introduces a lightweight token-merging strategy that combines similar tokens in vision transformers without retraining. LLaVA-PruMerge~\cite{shang2025llavaprumerge} adapts token reduction to large multimodal models by selecting important visual tokens and supplementing them with clustered token information. VisionZip~\cite{yang2025visionzip} keeps informative visual tokens and aggregates contextual tokens to form a compact visual sequence. DiViCo~\cite{wang2026divico} further studies visual token compression from a disentangled perspective, aiming to preserve useful visual information while reducing redundant tokens. These methods show that visual representations contain strong redundancy, but their redundancy estimates are still mainly derived from visual features or attention responses.

Video token compression must additionally handle temporal redundancy. Video-XL~\cite{shu2025videoxl} introduces visual summarization tokens to support hour-scale video understanding. Dynamic-VLM~\cite{wang2025dynamicvlm} explores dynamic visual token compression for VLLMs. PruneVid~\cite{huang2025prunevid} first removes relatively static and spatially close tokens, and then applies question-guided pruning. DyCoke~\cite{tao2025dycoke} combines temporal token compression with dynamic KV-cache compression to accelerate video large language models. FastVID~\cite{shen2025fastvid} adjusts pruning according to temporal segments and content density. FlashVID~\cite{fan2026flashvid} uses attention- and diversity-based token selection followed by tree-based spatiotemporal token merging. These methods explicitly consider video redundancy and have achieved strong efficiency-performance tradeoffs.

Despite these advances, existing methods share a common limitation: the candidate space for token selection or merging is still mainly constructed from visual tokens themselves. For intra-frame selection, this often requires evaluating importance, diversity, or distance over a large set of visual tokens. For temporal merging, global cross-frame matching introduces many pairwise feature comparisons, whereas fixed-position or overly local matching can miss correspondences after motion. Thus, visual token compression reduces the sequence sent to the language model, but the compression decision itself may still require substantial visual search. In contrast, this work uses coding information already available in the compressed video to restrict the candidate space before visual-feature evaluation.

\subsection{Video Codec-Guided Video Understanding}
\label{sec:codec_guided_video_understanding}

Recent video understanding methods have started to exploit video coding information
as a structural prior for visual representation. OneVision-Encoder~\cite{tang2026onevision} introduces
codec patchification, where motion magnitude and residual energy are used to
select informative patches before visual encoding, and a codec-aligned vision
encoder is trained to process the resulting sparse patch layout.
CoPE-VideoLM~\cite{sarkar2026cope} further builds a codec-aware token sequence for video language
models: I-frames are encoded by an image encoder, while MVs and residuals from
P-frames are encoded by lightweight transformer modules into compact
motion-residual tokens. These methods show that
codec primitives can expose temporal redundancy and local changes, but they
use such primitives to construct new visual representations before or during
model encoding.

More recent video MLLMs redesign the visual interface around codec structure.
LLaVA-OneVision-2~\cite{an2026llava} uses codec-stream tokenization, where bit-cost dynamics
determine adaptive temporal groups and motion-residual cues select salient
spatial evidence into compact visual canvases. AdaCodec~\cite{hou2026adacodec}
formulates video input as a predictive visual code: frames with high predictive
cost are represented by full visual tokens, while predictable intermediate
frames are represented by compact P-tokens derived from motion and residual
signals. Mage-VL~\cite{yang2026mage} develops a codec-native streaming
foundation model, in which motion vectors and residual energy guide the
selection of dynamic patches across anchor and predicted frames for efficient
online perception. In these systems, codec
information is embedded into the tokenization or visual encoding pipeline, and
the model is trained or fine-tuned to consume the resulting codec-native
representations.

Our work differs from these codec-native representation methods. We do not
introduce codec-derived visual tokens, train a codec encoder, or redesign the
visual interface of the video LLM. Instead, the video is still processed by the
original sampling and vision-encoding pipeline, and codec information is used
only after visual tokens have been produced. In this setting, reference-frame
dependencies, MVs, and residuals provide training-free guidance for token
compression over existing visual tokens. Our design enables codec-guided token
reduction without modifying the base model or requiring additional multimodal
training.

\begin{figure*}[t]
    \centering
    \includegraphics[width=\linewidth]{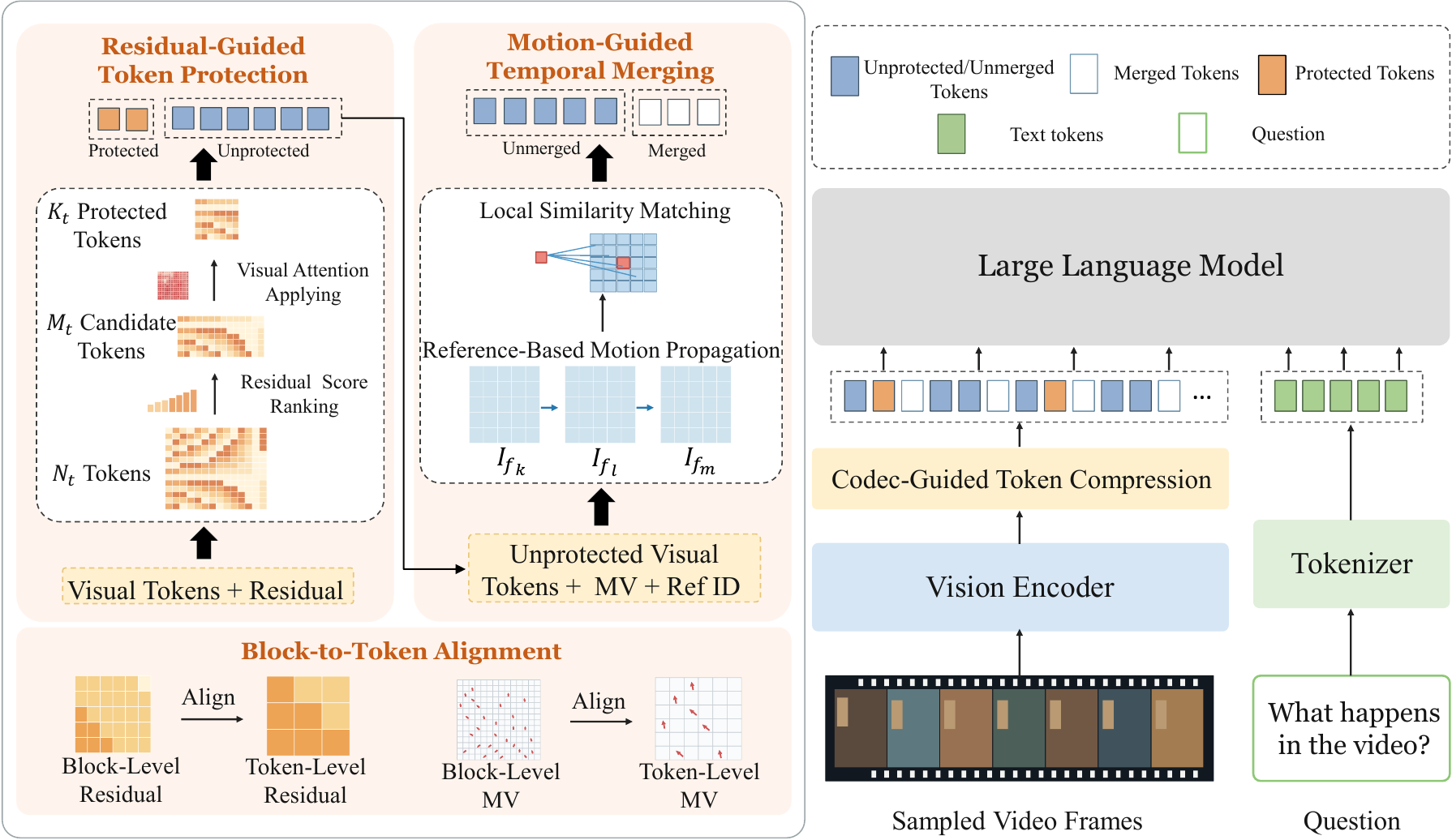}
    \caption{Overview of our proposed video codec-guided training-free token compression framework for video large language models.
    }
    \label{fig:framework_overview}
\end{figure*}

\section{Overview}
\label{sec:overview}

Video inputs to VLLMs are usually stored or transmitted as compressed bitstreams. A standard inference pipeline first decodes the bitstream into a frame sequence $V=\{I_f\}_{f=1}^{F}$, samples $T$ frames $\{I_{\tau_t}\}_{t=1}^{T}$ from this sequence, and feeds only the sampled frames into the vision encoder. In this work, we additionally retain the coding information available during decoding, denoted as $Z=\{Z_f\}_{f=1}^{F}$, where $Z_f$ contains the reference-frame relations, motion vectors, and residuals associated with frame $f$. Thus, the vision encoder still processes only the sampled frames $\{I_{\tau_t}\}_{t=1}^{T}$, while token compression can use coding information from the complete decoded video. This design provides motion and local-change priors for compression decisions without increasing the number of visually encoded frames.

Visual tokens and coding information are misaligned in both time and space. Temporal misalignment means that visual tokens exist only at the sampled frames $\{I_{\tau_t}\}_{t=1}^{T}$, whereas reference-frame relations and MVs are defined over the complete frame sequence $\{1,\ldots,F\}$. The coding reference of a current sampled frame is often not the preceding sampled frame, but an intermediate frame between them. Spatial misalignment means that visual tokens correspond to regions on a regular visual grid, whereas MVs and residuals are originally defined on coding blocks or pixels. A visual token may cover multiple coding blocks, and these blocks may have different reference directions and MVs. Therefore, the method first aligns block-level coding information from the complete video with the visual token grid, while preserving the role of intermediate frames in the reference chain.

Let $\{\tau_t\}_{t=1}^{T}$ be the original frame indices of the $T$ sampled frames. The vision encoder produces visual tokens $X_t=E_{\mathrm{vis}}(I_{\tau_t})\in\mathbb{R}^{N_t\times d}$ for each sampled frame, where $N_t$ is the number of visual tokens and $d$ is the feature dimension. Visual token compression is performed before these features enter the language model. We define the retention ratio as $\rho=N_{\mathrm{out}}/N_{\mathrm{in}}$, where $N_{\mathrm{in}}$ and $N_{\mathrm{out}}$ are the total token counts before and after compression. Unsampled frames do not enter the vision encoder, but their coding information remains in $Z$ and is used for motion propagation between sampled frames.

As illustrated in Fig.~\ref{fig:framework_overview}, our proposed video codec-guided training-free token compression method has three stages. The first stage is block-to-token alignment. According to the positions of coding blocks in each frame, block-level reference-frame IDs, MVs, and residuals are mapped to the visual token grid. The aligned coding representation contains two types of information: token-level residuals on sampled frames, and token-level motion information on both sampled and intermediate frames. A token-level motion item includes the token-level displacement and its reference-frame ID. Residuals are used to decide which tokens should be examined for protection, while motion information is used to construct cross-frame matching candidates.

The second stage is residual-guided token protection, which is performed before temporal merging. High-residual regions often correspond to local changes that are not well explained by motion compensation, such as newly visible content, occlusion, or motion boundaries. If these regions enter temporal merging too early, they may be incorrectly aggregated with redundant tokens from other frames. Therefore, residual scores are used to form a protection candidate pool. Visual feature attention is then applied within this pool to select the final protected tokens. The selected tokens form $\mathcal H_t$, are directly retained, and bypass subsequent temporal merging. The remaining tokens form $\mathcal R_t$ and enter the redundancy merging stage.

The third stage is motion-guided temporal merging. For each token in $\mathcal R_t$, the method starts from the current sampled frame and propagates its position along the reference-frame relations and MVs through intermediate coded frames. The accumulated motion predicts a position on the token grid of the preceding sampled frame. A local candidate set is then constructed around this predicted position, and visual feature similarity is computed only within this set. If the best match satisfies the temporal merging threshold, a temporal edge is created, and temporal redundancy trees are gradually formed. The trees of unprotected tokens are aggregated, and the protected tokens and aggregated representations together form the compressed visual sequence.

The key principle is to separate candidate generation from visual verification. Coding information proposes smaller and more plausible candidate sets, while visual feature similarity comparison is used to make the final protection and merging decisions. In this way, the method uses low-cost motion and residual information already present in compressed video to reduce visual search, while preserving the semantic decisions made by visual tokens.

\section{Methodology}
\label{sec:method}

The proposed method uses motion and residual information from compressed video to achieve efficient visual token compression. Residual information constrains the candidate pool for intra-frame token protection, while motion information constrains the candidate region for cross-frame token merging. This section introduces block-to-token alignment, residual-guided token protection, and motion-guided temporal merging.

\subsection{Block-to-Token Alignment}
\label{sec:codec_alignment}

\begin{figure}[t]
    \centering
    \includegraphics[width=\linewidth]{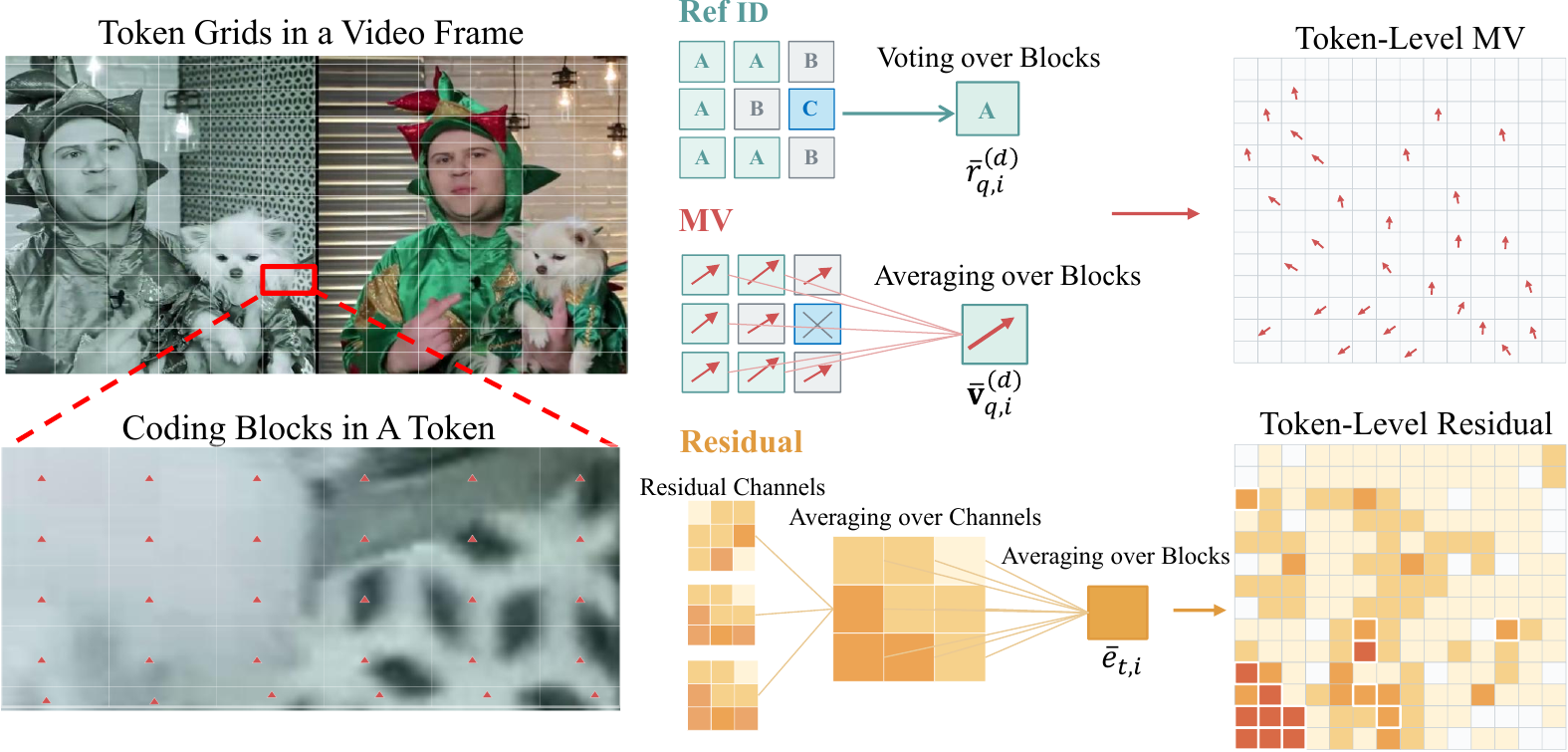}
    \caption{Illustration of the procedure for aligning block-level coding information to the visual token grid.}
    \label{fig:codec_token_alignment}
\end{figure}

Coding information and visual tokens use different spatial and temporal units. Coding information is defined over the complete video and is associated with coding blocks or pixels, whereas visual tokens are defined only on the sampled frames and follow the visual token grid. To use coding information for token compression, we align block-level reference-frame IDs, MVs, and pixel-level residuals with the visual token grid, as shown in Fig.~\ref{fig:codec_token_alignment}.

A token may cover multiple coding blocks, which may have different reference directions. For frame $f$, let $\mathcal B_{f,i}^{(\ell)}$ denote the set of coding blocks whose positions fall in token $i$ and that have a valid reference-frame ID and MV in reference direction $\ell$. Here, $\ell$ denotes the reference direction.
Since the coding blocks may refer to different frames, we first perform reference-frame ID voting within each direction. Specifically, we count how many blocks in $\mathcal B_{f,i}^{(\ell)}$  refer to each candidate reference frame ID $r$, and select the most frequent one as the token-level reference-frame ID:
\begin{equation}
\bar r_{f,i}^{(\ell)}
=
\underset{r}{\arg\max}
\sum_{b\in\mathcal B_{f,i}^{(\ell)}}
\mathbb I\!\left(r_{f,b}=r\right)
.
\label{eq:token_reference}
\end{equation}
Here, $\bar r_{f,i}^{(\ell)}$ is the voted token-level reference-frame ID, $r_{f,b}$ is the reference-frame ID of coding block $b$, and $\mathbb I(\cdot)$ equals 1 when its condition is true and 0 otherwise. This voting step avoids directly mixing MVs that point to different reference frames.

After the token-level reference-frame ID is determined, only the blocks supporting this reference-frame ID are used to compute the token-level MV. The token-level MV $\bar{\mathbf v}_{f,i}^{(\ell)}$ is then computed as
\begin{equation}
\bar{\mathbf v}_{f,i}^{(\ell)}
=
\frac{
\sum_{b\in\mathcal B_{f,i}^{(\ell)}}
\mathbb I\!\left(r_{f,b}=\bar r_{f,i}^{(\ell)}\right)
\mathbf v_{f,b}
}{
\sum_{b\in\mathcal B_{f,i}^{(\ell)}}
\mathbb I\!\left(r_{f,b}=\bar r_{f,i}^{(\ell)}\right)
}.
\label{eq:token_mv}
\end{equation}
where $\mathbf v_{f,b}$ denotes the MV of coding block $b$ after conversion to token-grid units. Thus, token-level motion information consists of the voted reference-frame ID $\bar r_{f,i}^{(\ell)}$ and the corresponding mean displacement $\bar{\mathbf v}_{f,i}^{(\ell)}$. The reference-frame ID determines which coded frame is visited next during reference-chain propagation, while the MV determines how the token position is updated.

Residual alignment is independent of reference-frame ID voting and MV aggregation. A residual describes the local difference that remains after motion compensation. It can therefore indicate motion boundaries, occlusion, newly visible content, and other regions with large prediction errors. Let $e_{\tau_t,c}(x)$ be the residual value at pixel position $x$ and pixel channel $c$ of the sampled frame $I_{\tau_t}$, and let $\Omega_{t,i}$ denote the set of pixels covered by token $i$ in $I_{\tau_t}$. The token-level residual score is computed by averaging the absolute residual values over pixel channels and over the token region:
\begin{equation}
\bar e_{t,i}
=
\frac{1}{|\Omega_{t,i}|}
\sum_{x\in\Omega_{t,i}}
\frac{1}{C}
\sum_{c=1}^{C}
\left|e_{\tau_t,c}(x)\right|.
\label{eq:token_residual}
\end{equation}
Here, $C$ is the number of pixel channels, and $|\Omega_{t,i}|$ is the number of pixels in the token region. The absolute value prevents positive and negative residuals from cancelling across channels.
\begin{figure}[t]
    \centering
    \includegraphics[width=\linewidth]{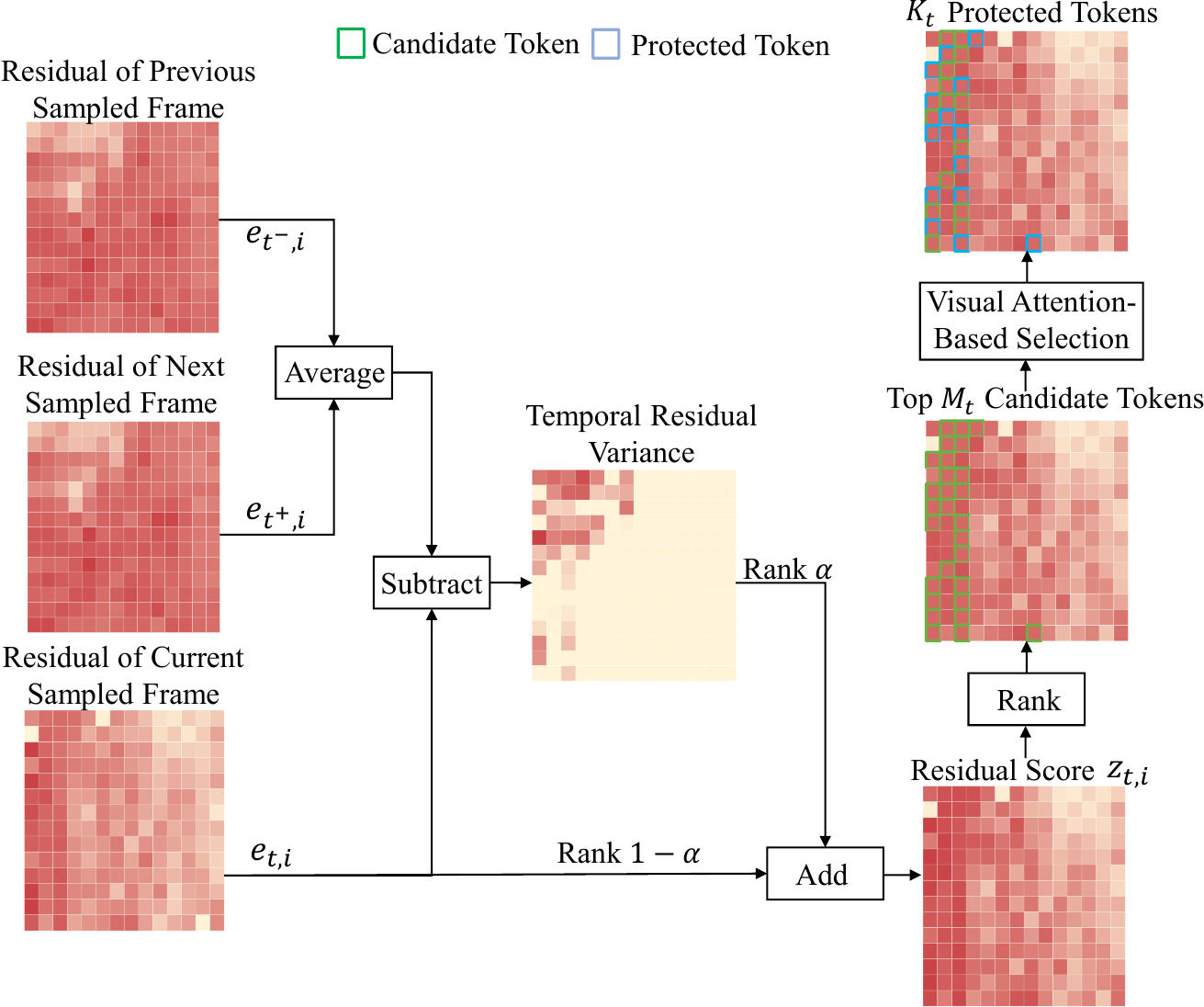}
    \caption{Illustration of the residual-guided token protection method.}
    \label{fig:residual_protection}
\end{figure}

\subsection{Residual-Guided Token Protection}
\label{sec:residual}

Large residuals often indicate local changes that are not well explained by motion compensation, such as newly visible content, occlusion, or motion boundaries. If tokens from these regions enter temporal merging too early, they may be incorrectly aggregated with redundant tokens from other frames. However, large residuals may also come from texture, compression noise, or coding errors. We therefore use residuals to construct a protection candidate pool, and then apply visual feature attention-based selection within the pool.

Using only the residual magnitude of the current frame may favor regions that remain difficult to predict throughout the video. To emphasize newly emerging or suddenly strengthened changes, as shown in Fig.~\ref{fig:residual_protection}, we define a positive temporal residual increase. Let $t^{-}=\max(1,t-1)$ and $t^{+}=\min(T,t+1)$. The positive temporal residual increase is
\begin{equation}
\delta_{t,i}
=
\left[
\bar e_{t,i}
-
\frac{
\bar e_{t^{-},i}+\bar e_{t^{+},i}
}{2}
\right]_{+},
\label{eq:residual_change}
\end{equation}
where $[z]_{+}=\max(z,0)$. This term compares residual changes at the same token-grid position and gives higher scores to token regions whose prediction error increases relative to adjacent sampled frames.

We rank-normalize the residual magnitude and the positive temporal residual increase within each sampled frame. Let $\rho_{t,i}^{e}$ and $\rho_{t,i}^{\delta}$ denote the normalized $\bar e_{t,i}$ and $\delta_{t,i}$, respectively, scaled to $[0,1]$. The weighted residual score is
\begin{equation}
z_{t,i}
=
(1-\alpha)\rho_{t,i}^{e}
+
\alpha\rho_{t,i}^{\delta},
\label{eq:combined_residual_score}
\end{equation}
where $\alpha\in[0,1]$ controls the contribution of the positive temporal residual increase. Given pool size $M_t$, the protection candidate pool is selected as
\begin{equation}
\mathcal P_t
=
\underset{\substack{\mathcal S\subseteq\{1,\ldots,N_t\}\\|\mathcal S|=M_t}}
{\arg\max}
\sum_{i\in\mathcal S}z_{t,i}.
\label{eq:residual_candidate_pool}
\end{equation}
Here, $\mathcal P_t$ contains the $M_t$ tokens with the largest combined residual scores. We set $K_t\leq M_t\leq N_t$, where $K_t$ is the final protection budget.

Within $\mathcal P_t$, we apply attention-based visual selection. Pairwise distances are computed inside the pool and calibrated by visual feature attention~\cite{fan2026flashvid}. Let $D_{ij}^{(t)}$ denote the calibrated token distance, and let $\mathcal H_t$ denote the current selected-token set. Subsequent tokens are selected by the max--min criterion:
\begin{equation}
i^{*}
=
\underset{i\in\mathcal P_t\setminus\mathcal H_t}{\arg\max}
\;
\underset{j\in\mathcal H_t}{\min}
D_{ij}^{(t)}.
\label{eq:semantic_selection}
\end{equation}
This criterion selects the candidate that is farthest from its closest selected token, encouraging diversity among protected tokens. Selection stops when $K_t$ tokens have been chosen, and the resulting set $\mathcal H_t$ is used as the protected-token set. The protected tokens bypass temporal merging and are directly retained. The remaining tokens form
\begin{equation}
\mathcal R_t=\{1,\ldots,N_t\}\setminus\mathcal H_t,
\end{equation}
and enter the motion-guided temporal merging stage. Since pairwise distances are computed only within $\mathcal P_t$, the intra-frame selection matrix is reduced from $N_t\times N_t$ to $M_t\times M_t$.
\begin{figure}[t]
    \centering
    \includegraphics[width=\linewidth]{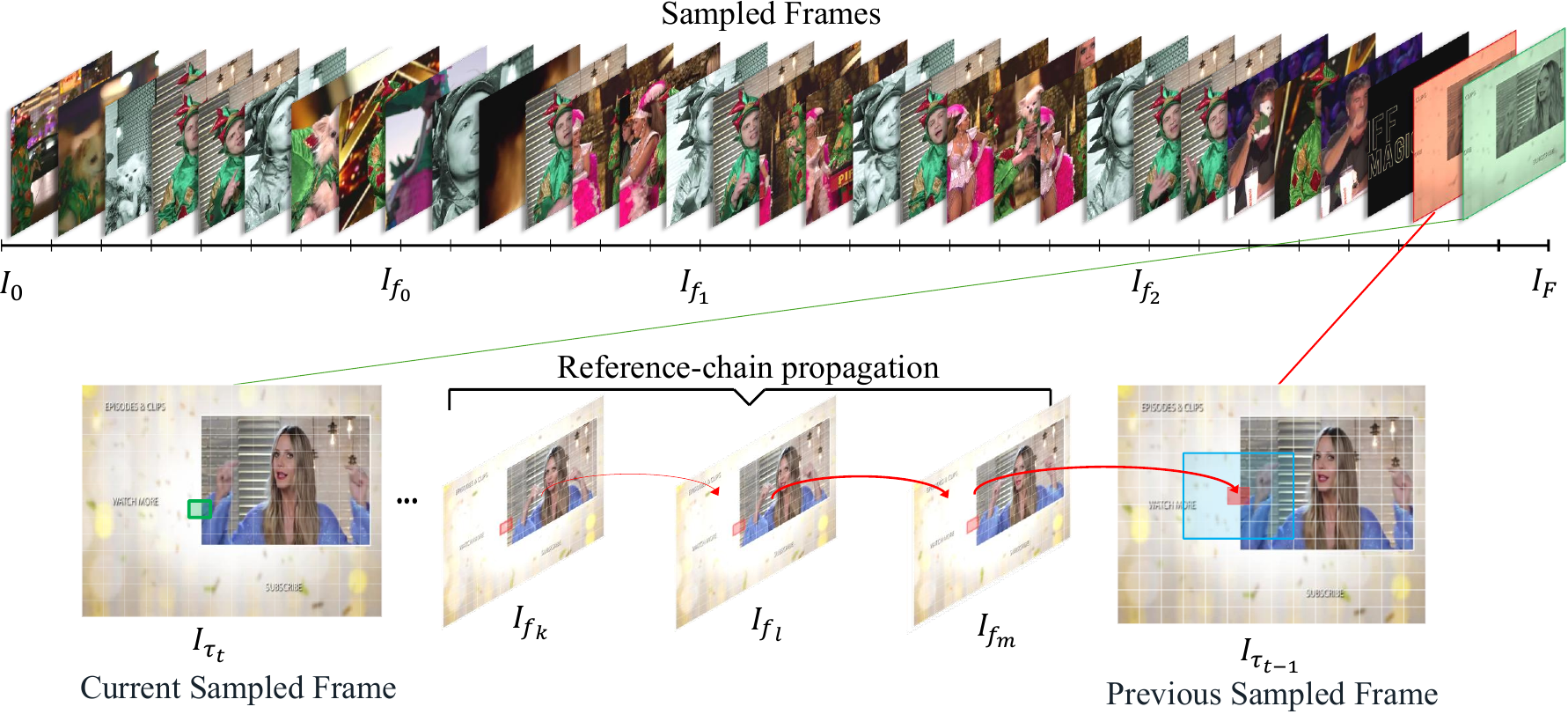}
    \caption{Illustration of motion-guided temporal merging through reference-chain propagation.}
    \label{fig:reference_chain}
\end{figure}
\subsection{Motion-Guided Temporal Merging}
\label{sec:motion}

This stage merges redundant tokens in $\mathcal R_t$ with the help of token-level MVs. A token-level MV describes the displacement from the current token position to its coding reference frame. However, the coding reference frame is usually an intermediate frame rather than the preceding sampled frame. Therefore, a single MV cannot directly establish token correspondence between two sampled frames. We address this mismatch by propagating token positions along the reference-frame chain in the complete video, thereby composing multiple short-range coding dependencies into a sampled-frame-level correspondence relation.

Let $I_{\tau_{t-1}}$ be the previous sampled frame and $I_{\tau_t}$ be the current sampled frame. For token $i\in\mathcal R_t$ in the current sampled frame, propagation starts from its token center $\mathbf p_i^{(0)}$ in $I_{\tau_t}$. At the $k$-th propagation step, the token is located in coded frame $I_{f_c^k}$, and we use the token-level reference-frame ID and MV whose reference direction moves toward $I_{\tau_{t-1}}$. The corresponding reference frame is $I_{f_r^k}$, and the token-level MV is $\bar{\mathbf v}_{f_c^k,i}$. The propagation rule is
\begin{equation}
\left(I_{f_c^{k+1}},\mathbf p_i^{(k+1)}\right)
=
\begin{cases}
\left(I_{\tau_{t-1}},
\mathbf p_i^{(k)}+\gamma_k\bar{\mathbf v}_{f_c^k,i}\right),
~ f_r^k \leq \tau_{t-1},\\[3pt]
\left(I_{f_r^k},
\mathbf p_i^{(k)}+\bar{\mathbf v}_{f_c^k,i}\right),
~ \tau_{t-1} < f_r^k < f_c^k.
\end{cases}
\label{eq:reference_chain}
\end{equation}

The reference-chain propagation between two adjacent sampled frames has two normal cases. If the next reference frame crosses the previous sampled frame, the MV from $I_{f_c^k}$ to $I_{f_r^k}$ is scaled to project the token position onto $I_{\tau_{t-1}}$, and propagation terminates. Otherwise, if the reference frame remains between the current coded frame and the previous sampled frame, the token position is updated to the next reference frame, and propagation continues along the reference-frame chain. For the crossing case, the temporal scaling factor is defined as
\begin{equation}
\gamma_k=
\frac{f_c^k-\tau_{t-1}}{f_c^k-f_r^k},
\qquad
0<\gamma_k\leq\gamma_{\max}.
\label{eq:motion_scaling}
\end{equation}
Here, $\gamma_{\max}$ prevents the MV from being scaled under an unreliable temporal span. Terminal projection is allowed only when the reference frame crosses the previous sampled frame and the temporal gap between the current coded frame and the previous sampled frame is no larger than $\eta$.

Reference-chain propagation may encounter missing or abnormal motion information. If the token cell at the current coded frame $I_{f_c^k}$ has no valid token-level MV and reference-frame ID, we first search nearby token cells within a radius of 1 for a valid pair. When multiple valid pairs are found, the nearest one is used. If no valid pair is available in the neighborhood, we reuse the MV from the previous propagation step to update the token position. 

If propagation terminates at step $K_i$, the final propagated position is denoted by
\begin{equation}
\mathbf p_{\tau_t\rightarrow\tau_{t-1},i}
=
\mathbf p_i^{(K_i)}.
\label{eq:final_propagated_position}
\end{equation}
This position lies on the token grid of the previous sampled frame $I_{\tau_{t-1}}$. Let $\mathbf p_{\tau_{t-1},j}$ denote the grid center of token $j$ in $I_{\tau_{t-1}}$. For a normally terminated propagation chain, the local candidate set with search radius $r$ is
\begin{equation}
\mathcal C_{t,i}^{\mathrm{loc}}(r)
=
\left\{
j\in\mathcal R_{t-1}:
\left\|
\mathbf p_{\tau_{t-1},j}
-
\mathbf p_{\tau_t\rightarrow\tau_{t-1},i}
\right\|_{\infty}
\leq r
\right\}.
\label{eq:local_candidates}
\end{equation}

The actual candidate set used for visual matching is denoted by $\mathcal C_{t,i}$. For a normally terminated chain, we set $\mathcal C_{t,i}=\mathcal C_{t,i}^{\mathrm{loc}}(r)$. If propagation terminates abnormally, the same construction is applied with a wider relaxed radius $r^*$. If the resulting candidate set contains fewer than a prescribed minimum number of tokens, we set $\mathcal C_{t,i}=\mathcal R_{t-1}$ and fall back to global matching over the unprotected tokens in the previous sampled frame.

Visual similarity is evaluated only within $\mathcal C_{t,i}$. Since $X_t\in\mathbb R^{N_t\times d}$ denotes the visual tokens of sampled frame $I_{\tau_t}$, let $\mathbf x_{t,i}$ and $\mathbf x_{t-1,j}$ denote the corresponding token features in $X_t$ and $X_{t-1}$. The best matched token and its similarity are
\begin{equation}
\begin{aligned}
j_{t,i}^{*}
&=
\underset{j\in\mathcal C_{t,i}}{\arg\max}\;
\frac{\mathbf x_{t,i}^{\mathsf T}\mathbf x_{t-1,j}}
{\|\mathbf x_{t,i}\|_2\|\mathbf x_{t-1,j}\|_2},\\
s_{t,i}^{*}
&=
\frac{\mathbf x_{t,i}^{\mathsf T}\mathbf x_{t-1,j_{t,i}^{*}}}
{\|\mathbf x_{t,i}\|_2\|\mathbf x_{t-1,j_{t,i}^{*}}\|_2}.
\end{aligned}
\label{eq:local_semantic_matching}
\end{equation}

The temporal merge decision is
\begin{equation}
\mu_{t,i}
=
\mathbb I\!\left(s_{t,i}^{*}\geq\theta_{\mathrm{tem}}\right),
\qquad i\in \mathcal R_t.
\label{eq:temporal_merge_decision}
\end{equation}
where $\theta_{\mathrm{tem}}$ is the temporal merging threshold. When $\mu_{t,i}=1$, token $i$ is merged with $j_{t,i}^{*}$. Otherwise, token $i$ is not merged with the previous sampled frame.



\section{Experiments}
\label{sec:experiments}

\subsection{Experimental Setup}
\label{sec:settings}

\subsubsection{Benchmarks}
We evaluate the proposed method on four video-understanding benchmarks: Video-MME~\cite{fu2025videomme}, EgoSchema~\cite{mangalam2023egoschema}, LongVideoBench~\cite{wu2024longvideobench}, and MVBench~\cite{li2024mvbench}. These benchmarks cover complementary aspects of video understanding. Video-MME contains short, medium, and long videos, allowing us to examine performance under different temporal spans. EgoSchema focuses on long-form egocentric video reasoning. LongVideoBench evaluates retrieval and reasoning over extended temporal contexts. MVBench covers a broad set of temporal perception tasks. We report the short, medium, and long subsets of Video-MME, both the Subset and Total splits of EgoSchema, and the overall scores of LongVideoBench and MVBench.
\begin{table*}[htb]
\centering\small
\caption{Comparison with existing training-free token compression methods on LLaVA-OneVision and LLaVA-Video.}
\label{tab:main_results}
\setlength{\tabcolsep}{2.5pt}
\renewcommand{\arraystretch}{0.94}
\begin{adjustbox}{max width=\textwidth,center}
\begin{tabular}{l c c c c c c c c c c c}
\toprule
\multirow{2}{*}{Method} & \multirow{2}{*}{$\rho$} & \multicolumn{4}{c}{Video-MME} & \multicolumn{2}{c}{EgoSchema} & \multirow{2}{*}{LongVideoBench} & \multirow{2}{*}{MVBench} & \multicolumn{2}{c}{Average} \\
\cmidrule(lr){3-6}\cmidrule(lr){7-8}\cmidrule(l){11-12}
 & & Short & Medium & Long & Overall & Subset & Total & & & Score & Relative Accuracy (\%) \\
\midrule
\multicolumn{12}{c}{\emph{LLaVA-OneVision}} \\
\midrule
Vanilla   & 100\% & 69.9 & 56.7 & 48.9 & 58.5 & 62.2 & 60.3 & 56.6 & 58.3 & 58.4 & 100.0 \\
\midrule
FastV     & 25\% & 68.1 & 54.7 & 46.8 & 56.5 & 60.4 & 57.8 & 55.4 & 56.4 & 56.5 & 96.7 \\
VisionZip & 25\% & 68.8 & \textbf{57.3} & 48.2 & 58.1 & 63.0 & 60.4 & 56.4 & 57.8 & 58.2 & 99.7 \\
PruneVID  & 25\% & 67.3 & 54.8 & 47.2 & 56.4 & 61.0 & 58.1 & 55.4 & 56.8 & 56.7 & 97.1 \\
FastVID   & 25\% & 69.9 & 56.3 & 47.4 & 57.9 & 61.2 & 59.5 & 55.9 & 58.1 & 57.8 & 99.0 \\
FlashVID  & 25\% & 70.2 & 55.3 & 48.9 & 58.1 & \textbf{64.2} & \textbf{61.3} & 56.9 & \textbf{58.3} & \textbf{58.7} & \textbf{100.5} \\
Ours      & 25\% & \textbf{70.4} & 56.8 & \textbf{49.8} & \textbf{59.0} & 63.6 & \textbf{61.3} & \textbf{57.0} & 57.4 & \textbf{58.7} & \textbf{100.5} \\
\midrule
FastV     & 20\% & 66.3 & 53.9 & 46.9 & 55.7 & 60.6 & 57.6 & \textbf{56.0} & 56.0 & 56.3 & 96.4 \\
VisionZip & 20\% & 68.6 & \textbf{57.0} & 48.3 & 58.0 & 62.0 & 60.0 & 55.4 & 57.6 & 57.7 & 98.8 \\
PruneVID  & 20\% & 67.2 & 53.9 & 48.2 & 56.4 & 63.2 & 60.2 & 55.2 & 56.2 & 57.0 & 97.6 \\
FastVID   & 20\% & 69.9 & 56.3 & 47.4 & 57.9 & 61.2 & 59.5 & 55.9 & \textbf{58.1} & \textbf{57.9} & \textbf{99.1} \\
FlashVID  & 20\% & 70.4 & 55.3 & 48.2 & 58.0 & \textbf{65.8} & 60.3 & 55.4 & 58.0 & \textbf{57.9} & \textbf{99.1} \\
Ours      & 20\% & \textbf{70.7} & 55.6 & \textbf{48.8} & \textbf{58.3} & 62.6 & \textbf{60.9} & 55.1 & 56.8 & 57.8 & 99.0 \\
\midrule
FastV     & 15\% & 64.6 & 54.0 & 45.3 & 54.6 & 59.8 & 56.6 & 54.8 & 55.0 & 55.2 & 94.5 \\
VisionZip & 15\% & 63.8 & 54.6 & 48.3 & 55.6 & 62.8 & 60.0 & 54.1 & 53.5 & 55.8 & 95.5 \\
FastVID   & 15\% & 69.7 & \textbf{55.8} & 47.7 & \textbf{57.7} & 58.8 & 58.9 & \textbf{56.7} & \textbf{58.2} & 57.9 & 99.1 \\
PruneVID  & 15\% & 67.2 & 52.8 & 46.7 & 56.1 & 61.6 & 57.7 & 54.5 & 55.1 & 55.7 & 95.4 \\
FlashVID  & 15\% & \textbf{70.3} & 54.5 & 47.8 & 57.5 & \textbf{64.6} & 60.2 & 56.5 & 57.9 & \textbf{58.0} & \textbf{99.3} \\
Ours      & 15\% & 68.4 & 53.2 & \textbf{49.2} & 57.0 & 62.4 & \textbf{60.3} & 56.3 & 57.1 & 57.7 & 98.8 \\
\midrule
FastV     & 10\% & 60.9 & 52.2 & 44.9 & 52.7 & 59.0 & 56.0 & 52.4 & 53.4 & 53.6 & 91.8 \\
VisionZip & 10\% & 60.3 & 52.9 & 46.7 & 53.3 & 61.6 & 58.5 & 49.4 & 54.8 & 54.0 & 92.5 \\
PruneVID  & 10\% & 65.9 & 52.8 & 45.6 & 54.7 & 60.0 & 57.2 & 54.0 & 53.7 & 54.9 & 94.0 \\
FastVID   & 10\% & 68.1 & \textbf{55.7} & 47.8 & 57.2 & 58.8 & 58.7 & \textbf{55.7} & 57.0 & 57.1 & 97.8 \\
FlashVID  & 10\% & \textbf{69.7} & 54.7 & 47.9 & \textbf{57.4} & \textbf{64.4} & 60.0 & 54.2 & \textbf{57.8} & \textbf{57.4} & \textbf{98.3} \\
Ours      & 10\% & 67.0 & 53.1 & \textbf{49.9} & 56.7 & 63.0 & \textbf{60.7} & 54.2 & 56.3 & 57.0 & 97.6 \\
\midrule
\multicolumn{12}{c}{\emph{LLaVA-Video}} \\
\midrule
Vanilla   & 100\% & 77.0 & 62.1 & 53.3 & 64.2 & 59.4 & 57.3 & 59.5 & 61.9 & 60.7 & 100.0 \\
\midrule
FastV     & 20\% & 69.3 & 58.3 & 49.9 & 59.2 & 54.8 & 54.1 & 56.0 & 58.4 & 56.9 & 93.7 \\
VisionZip & 20\% & 72.3 & 59.6 & \textbf{53.3} & 61.7 & 59.0 & 56.4 & 58.0 & 59.8 & 59.0 & 97.2 \\
FastVID   & 20\% & \textbf{74.6} & \textbf{60.8} & 52.3 & \textbf{62.6} & 57.0 & 55.0 & 57.1 & 60.2 & 58.7 & 96.7 \\
FlashVID  & 20\% & 74.3 & 59.8 & 51.2 & 61.8 & \textbf{67.6} & \textbf{60.7} & 59.0 & \textbf{60.4} & \textbf{60.5} & \textbf{99.7} \\
Ours      & 20\% & 73.3 & 59.4 & 51.6 & 61.4 & 64.8 & 60.6 & \textbf{59.4} & 59.9 & 60.3 & 99.3 \\
\midrule
FastV     & 10\% & 64.3 & 53.8 & 49.2 & 55.8 & 50.6 & 51.1 & 53.6 & 56.2 & 54.2 & 89.3 \\
VisionZip & 10\% & 69.4 & 57.9 & 51.2 & 59.5 & 54.4 & 53.9 & 54.5 & 58.5 & 56.6 & 93.2 \\
FastVID   & 10\% & 71.8 & 57.3 & 50.2 & 59.8 & 54.8 & 52.4 & 56.9 & 59.3 & 57.1 & 94.1 \\
FlashVID  & 10\% & \textbf{71.9} & \textbf{58.8} & \textbf{51.3} & \textbf{60.7} & 65.2 & \textbf{59.5} & \textbf{58.0} & \textbf{59.7} & \textbf{59.5} & \textbf{98.0} \\
Ours      & 10\% & 71.6 & 57.7 & 51.0 & 60.1 & \textbf{65.4} & 59.3 & 56.7 & 58.8 & 58.7 & 96.8 \\
\bottomrule
\end{tabular}
\end{adjustbox}
\end{table*}


\begin{figure*}[t]
    \centering
    \includegraphics[width=0.9\linewidth]{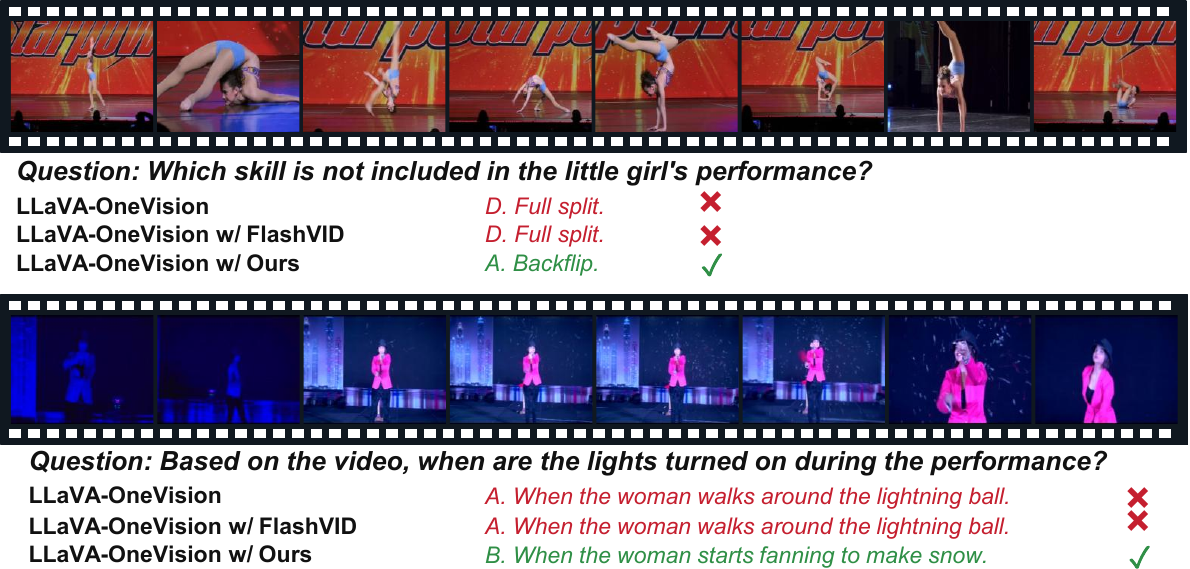}
    \caption{Subjective comparison results between FlashVID and our method.}
    \label{fig:subjective}
\end{figure*}

\begin{table*}[htb]
\centering\small
\caption{Comparison with existing training-free token compression methods on Qwen2.5-VL.}
\label{tab:qwen_results}
\setlength{\tabcolsep}{2.5pt}
\renewcommand{\arraystretch}{0.94}
\begin{adjustbox}{max width=\textwidth,center}
\begin{tabular}{lcccccccccc}
\toprule
Method & $\rho$ & \multicolumn{4}{c}{Video-MME} & \multicolumn{2}{c}{EgoSchema} & LongVideoBench & MVBench & Average Score \\
\cmidrule(lr){3-6}\cmidrule(lr){7-8}
 & & Short & Medium & Long & Overall & Subset & Total & & & \\
\midrule
Vanilla & 100\% & 72.6 & 61.4 & 49.9 & 61.3 & 60.2 & 58.3 & 58.9 & 68.0 & 61.6 \\
\midrule
FastV & 20\% & 69.4 & 57.0 & \textbf{51.2} & 59.2 & 60.2 & \textbf{57.1} & 54.2 & \textbf{66.8} & 59.3 \\
VisionZip & 20\% & 69.6 & 57.2 & 50.2 & 59.0 & 58.6 & 56.6 & 56.3 & 66.4 & \textbf{59.6} \\
FastVID & 20\% & 69.9 & 56.3 & 49.7 & 58.6 & 57.2 & 56.4 & \textbf{57.8} & 64.7 & 59.4 \\
FlashVID & 20\% & \textbf{70.6} & \textbf{59.0} & 49.4 & \textbf{59.7} & 58.8 & 55.8 & 53.9 & 65.1 & 58.6 \\
Ours & 20\% & 70.3 & 58.4 & 50.1 & 59.6 & \textbf{61.2} & 56.3 & 53.6 & 64.9 & 58.6 \\
\midrule
FastV & 10\% & 63.7 & \textbf{54.7} & \textbf{49.3} & 55.9 & \textbf{58.6} & 54.9 & 51.1 & 63.6 & 57.3 \\
VisionZip & 10\% & 67.0 & \textbf{54.7} & 47.6 & 56.4 & 55.8 & 55.5 & 54.5 & \textbf{64.3} & \textbf{57.7} \\
FastVID & 10\% & 66.3 & 53.6 & 49.0 & 56.3 & 56.0 & \textbf{55.6} & \textbf{55.4} & 62.3 & 57.4 \\
FlashVID & 10\% & 68.7 & \textbf{54.7} & 48.4 & 57.3 & \textbf{58.6} & 54.7 & 53.0 & 62.1 & 56.8 \\
Ours & 10\% & \textbf{69.2} & 54.2 & 49.1 & \textbf{57.5} & 58.4 & 55.1 & 52.8 & 62.4 & 57.0 \\
\bottomrule
\end{tabular}
\end{adjustbox}
\end{table*}

\subsubsection{Evaluated Video Large Language Models}
We conduct experiments on three representative VLLMs: LLaVA-OneVision~\cite{li2024llavaonevision}, LLaVA-Video~\cite{zhang2024llavavideo}, and Qwen2.5-VL-7B-Instruct~\cite{bai2025qwen25vl}. LLaVA-OneVision and LLaVA-Video use related model designs but different video sampling lengths, while Qwen2.5-VL adopts a different visual processing pipeline and a dynamic visual grid. Evaluating these models allows us to test whether codec-guided candidate construction remains effective under different numbers and spatial layouts of visual tokens.

\subsubsection{Baselines}
We compare with the uncompressed model and five training-free visual token compression methods. FastV~\cite{chen2024fastv} represents attention-guided visual token pruning. VisionZip~\cite{yang2025visionzip} selects informative visual tokens and aggregates contextual tokens. PruneVID~\cite{huang2025prunevid} and FastVID~\cite{shen2025fastvid} are video-oriented pruning methods. FlashVID~\cite{fan2026flashvid} is the state-of-the-art training-free video token compression method based on spatiotemporal token merging.
For each model, the average score is computed as the arithmetic mean of Video-MME Overall, EgoSchema Total, LongVideoBench, and MVBench. Relative accuracy is the ratio between this average score and that of the corresponding uncompressed model. This aggregation assigns the same weight to each benchmark and avoids bias toward benchmarks with more questions.



\subsubsection{Implementation Details}
We implement our proposed method based on FlashVID, which is the state-of-the-art training-free token compression method for VLLMs.  All comparisons between FlashVID and our method use the same base model, sampled frames, target retention ratio, and decoding procedure. Our method only modifies the candidate construction used for token protection and temporal matching. 
The local matching radius $r$ is set to 2, and the relaxed matching radius $r^*$ is set to 4. The maximum terminal scaling factor $\gamma_{\max}$  is 2, and the temporal merging threshold $\theta_{\mathrm{tem}}$ is 0.75. 
The temporal gap threshold $\eta$ in reference-chain propagation is 4. 
The positive temporal residual increase weight $\alpha$ is set to 0.2. The residual candidate pool is twice the target number of protected tokens in one frame. All experiments are conducted on an NVIDIA RTX PRO 6000 GPU with 96\,GB memory.

\subsection{Comparison with Baseline Methods}
\label{sec:main_results}

\subsubsection{Results on LLaVA-OneVision}
Table~\ref{tab:main_results} reports the results on LLaVA-OneVision under four retention ratios. We use the Average Score column to compare absolute performance, and use Relative Accuracy to measure the percentage of the Vanilla
average score retained after compression. Across all retention ratios, our method maintains an average score close to FlashVID. At $\rho=25\%$, both methods obtain an average score of 58.7, corresponding to 100.5\% relative accuracy. Even when the retention ratio is reduced to $\rho=10\%$, our method achieves an average score of 57.0 and retains 97.6\% of the Vanilla average score. These results indicate that the proposed video codec-guided token compression method preserves overall performance under different token budgets. Notably, our method consistently improves the Video-MME Long score over FlashVID, with gains of 0.9, 0.6, 1.4, and 2.0 points at $\rho=25\%$, $20\%$, $15\%$, and $10\%$, respectively. This suggests that codec-guided candidate construction is particularly useful when temporal correspondences are more difficult to identify from sparse sampled frames.

Fig.~\ref{fig:subjective} further shows two qualitative examples where our method gives the correct answer while both Vanilla and FlashVID fail. Although Vanilla retains all visual tokens, redundant or question-irrelevant information may distract the model from key evidence. FlashVID searches over a larger candidate space and may merge tokens that are visually similar but temporally
unrelated. In contrast, our method uses codec reference relationships to guide cross-frame token correspondence, leading to more reliable temporal merging while reducing feature comparisons.

\subsubsection{Results on LLaVA-Video}
Table~\ref{tab:main_results} also reports the results on LLaVA-Video. At $\rho=20\%$, our method obtains an average score of 60.3, only 0.2 points below FlashVID, and retains 99.3\% relative accuracy. It also achieves the best
LongVideoBench score among the compressed methods at this retention ratio. These results show that the proposed codec-guided token compression method remains effective on a different VLLM with a different frame sampling
setting.

At $\rho=10\%$, our method obtains an average score of 58.7 and retains 96.8\% relative accuracy. Although FlashVID has a higher average score, our method remains competitive on Video-MME and EgoSchema, and achieves the best EgoSchema Subset score among all compressed methods at this retention ratio. These results indicate that codec-guided candidate construction can be applied across different VLLMs and remains stable under strong compression.

\subsubsection{Results on Qwen2.5-VL}
Table~\ref{tab:qwen_results} reports the results on Qwen2.5-VL. This model uses a different visual processing pipeline and a dynamic visual grid, making the alignment between codec blocks and visual tokens more challenging. Compared with the direct baseline FlashVID, our method obtains the same average score of 58.6 at $\rho=20\%$ and improves the average score from 56.8 to 57.0 at $\rho=10\%$. On Video-MME, our method achieves comparable Overall scores and
consistently improves the Long subset by 0.7 points at both retention ratios. It also improves EgoSchema Subset at $\rho=20\%$ and EgoSchema Total at $\rho=10\%$. These results show that codec-guided candidate construction remains applicable to a VLLM with a dynamic visual token layout. 

\begin{table}[t]
\centering
\caption{Reduction of feature comparisons relative to FlashVID on Video-MME with LLaVA-OneVision. The Video-MME column reports our score, with the FlashVID score shown in parentheses.}
\scalebox{0.9}{
\begin{tabular}{cccc}
\toprule
$\rho$ & Selection Reduction (\%) & Temporal Reduction (\%) & Video-MME \\
\midrule
25\% & 79.8 & 64.7 & \textbf{59.0} (58.1) \\
20\% & 87.2 & 64.7 & 58.3 (58.0) \\
15\% & 93.0 & 64.7 & 57.0 (57.5) \\
10\% & \textbf{96.6} & 64.7 & 56.7 (57.4) \\
\bottomrule
\end{tabular}}
\label{tab:similarity_cost}
\end{table}

\subsection{Complexity Analysis}
\label{sec:complexity}

We analyze the feature comparisons required by intra-frame token protection and
cross-frame temporal matching. For token protection, FlashVID computes a
pairwise visual-distance matrix over all $N_t$ tokens in sampled frame
$I_{\tau_t}$. In our method, residual guidance restricts visual selection to the
candidate pool $\mathcal P_t$, where $|\mathcal P_t|=M_t$. Therefore, the
number of pairwise entries evaluated for token protection is reduced from
$N_t^2$ to $M_t^2$ for frame $t$, and from $\sum_{t=1}^{T}N_t^2$ to
$\sum_{t=1}^{T}M_t^2$ over the sampled video.

For temporal matching, exhaustive matching between the unprotected tokens in
two adjacent sampled frames would compare each token in $\mathcal R_t$ with all
tokens in $\mathcal R_{t-1}$. This requires
$|\mathcal R_t||\mathcal R_{t-1}|$ visual-similarity evaluations for frame pair
$(t-1,t)$. Motion-guided temporal merging instead evaluates visual similarity
only within the candidate set $\mathcal C_{t,i}$ constructed for each token
$i\in\mathcal R_t$. The total number of temporal matching comparisons is
therefore $\sum_{t=2}^{T}
\sum_{i\in\mathcal R_t}
|\mathcal C_{t,i}|$. 
For normally terminated propagation chains, $\mathcal C_{t,i}$ is a local
window centered at the propagated position. With search radius $r$, the number
of candidates is at most $(2r+1)^2$. If propagation terminates abnormally or
the local set is too small, the relaxed or global fallback rule determines
$\mathcal C_{t,i}$.

Table~\ref{tab:similarity_cost} reports the measured reduction in feature
comparisons relative to FlashVID on Video-MME with LLaVA-OneVision. The
temporal comparison reduction is 64.7\% across all evaluated retention ratios,
since the motion-guided search rule is unchanged across these settings. The
selection comparison reduction increases from 79.8\% at $\rho=25\%$ to 96.6\%
at $\rho=10\%$, because the residual candidate pool becomes smaller as the
token budget decreases. Meanwhile, the Video-MME scores remain close to
FlashVID under matched retention ratios. These results show that motion and
residual guidance reduce different comparison costs: motion guidance restricts
cross-frame temporal matching, while residual guidance restricts intra-frame
token protection.
\begin{table}[t]
\centering\small
\setlength{\tabcolsep}{2.5pt}
\renewcommand{\arraystretch}{0.95}
\caption{Effectiveness of motion and residual guidance on Video-MME with LLaVA-OneVision ($\rho=25\%$).}
\label{tab:module_ablation}
\begin{tabular}{lcccc}
\toprule
Method & Short & Medium & Long & Overall \\
\midrule
Motion-Guided Only & \textbf{71.4} & 56.5 & 48.9 & \textbf{59.0} \\
Residual-Guided Only & 70.3 & 55.8 & 49.0 & 58.4 \\
Combined & 70.4 & \textbf{56.8} & \textbf{49.8} & \textbf{59.0} \\
\bottomrule
\end{tabular}
\end{table}
\begin{table}[t]
\centering\small
\setlength{\tabcolsep}{3pt}
\renewcommand{\arraystretch}{0.95}
\caption{Effectiveness of temporal residual increase on Video-MME with LLaVA-OneVision ($\rho=25\%$).}
\label{tab:alpha_ablation}
\begin{tabular}{ccccc}
\toprule
$\alpha$ & Short & Medium & Long & Overall \\
\midrule
0 & \textbf{70.9} & 55.9 & 48.9 & 58.6 \\
0.1 & \textbf{70.9} & 55.4 & 49.3 & 58.6 \\
0.2 & 70.4 & \textbf{56.6} & \textbf{49.8} & \textbf{59.0} \\
\bottomrule
\end{tabular}
\end{table}
\subsection{Ablation Studies}
\label{sec:ablation}
\subsubsection{Effectiveness of Motion and Residual Guidance}
Table~\ref{tab:module_ablation} evaluates the two codec-guided components separately. Motion-guided temporal merging uses reference-chain propagation and
MVs to construct cross-frame matching candidates. Residual-guided token protection uses residual scores to form a compact candidate pool. The motion-guided-only variant achieves an Overall score of 59.0 and the best Short score, showing that codec-derived motion provides effective correspondence priors for temporal token merging. The residual-guided-only variant obtains a lower Overall score, but gives a similar Long score, suggesting that residual cues capture useful local evidence such as newly visible content, occlusion, and motion boundaries. In addition, residual guidance reduces the pairwise visual-distance computations used for token protection by limiting selection to a smaller candidate pool.
Combining the two components keeps the best Overall score of 59.0 and obtains the best Medium and Long scores. Compared with motion guidance alone, the
combined method slightly decreases the Short score but improves the Medium and Long scores, while also retaining the reduced protection cost introduced by the
residual candidate pool. This behavior is consistent with their complementary roles: motion guidance restricts cross-frame temporal matching, whereas residual guidance restricts intra-frame token protection and preserves locally
changing regions before merging. The combined design therefore improves the accuracy-complexity tradeoff without increasing the output-token budget.

\begin{table}[t]
\centering\small
\setlength{\tabcolsep}{3pt}
\renewcommand{\arraystretch}{0.95}
\caption{Effectiveness of the local search radius on Video-MME with LLaVA-OneVision ($\rho=25\%$).}
\label{tab:radius_ablation}
\begin{tabular}{ccccc}
\toprule
$r$ & Short & Medium & Long & Overall \\
\midrule
1 & 70.3 & 56.3 & 49.7 & 58.8 \\
2 & 70.4 & \textbf{56.8} & 49.8 & 59.0 \\
3 & 70.3 & 55.6 & \textbf{49.9} & 58.6 \\
4 & \textbf{70.9} & \textbf{56.8} & 49.8 & \textbf{59.1} \\
\bottomrule
\end{tabular}
\end{table}

\begin{figure}[t]
    \centering
    \includegraphics[width=\linewidth]{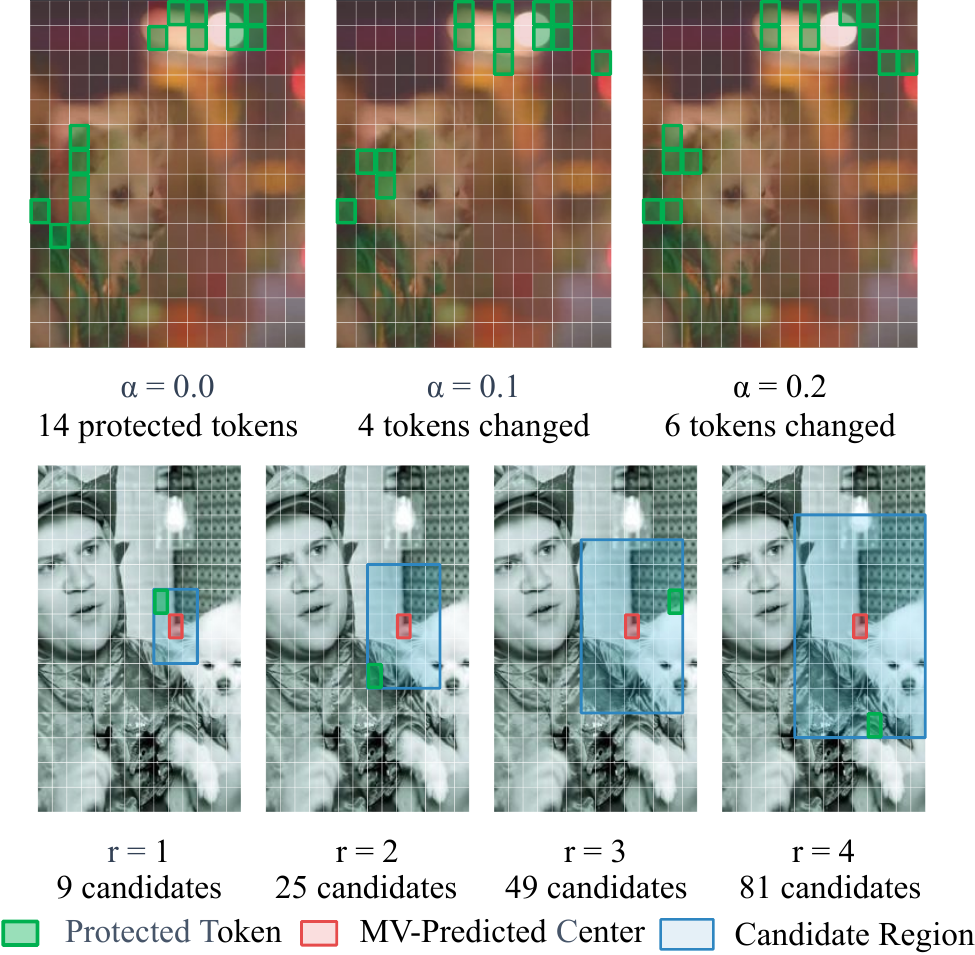}
\caption{Effectiveness of temporal residual increase and local search radius. The top part shows the protected tokens under different temporal residual
increase weights. Adjusting $\alpha$ changes the protected tokens while keeping the protection budget fixed. The bottom part shows the candidate regions for
local matching under different search radii $r$.}    \label{fig:parameter_visualization}
\end{figure}

\subsubsection{Effectiveness of Temporal Residual Increase}
We analyze the temporal residual increase used in residual-guided token protection. As shown in Fig.~\ref{fig:parameter_visualization}, changing
$\alpha$ changes which tokens are selected for protection while keeping the protection budget fixed. When $\alpha=0$, the residual candidate pool is determined only by the residual magnitude of the current frame. When
$\alpha>0$, the candidate pool additionally favors token locations whose residuals increase relative to adjacent sampled frames. This helps distinguish newly emerging changes from regions that remain difficult to predict throughout the video.  Table~\ref{tab:alpha_ablation} reports the performance under different values of $\alpha$. Adding temporal residual increase mainly benefits the Medium and
Long subsets. With $\alpha=0.2$, the Medium and Long scores increase to 56.6 and 49.8, respectively, and the Overall score improves from 58.6 to 59.0. This indicates that temporal residual increase helps identify local changes that are more relevant in longer temporal contexts. We therefore set $\alpha=0.2$ as the default value.

\subsubsection{Effectiveness of the Local Search Radius}
We further analyze the local search radius $r$. As shown in
Fig.~\ref{fig:parameter_visualization}, this parameter controls the tradeoff between candidate coverage and feature-comparison cost. A smaller radius relies
more strongly on the accuracy of motion propagation, while a larger radius can tolerate larger prediction errors at the cost of more candidate comparisons. Table~\ref{tab:radius_ablation} reports the performance under different local search radii. Increasing $r$ generally improves the Overall score, but the gain
becomes marginal when the radius is enlarged to 4. In particular, increasing $r$ from 2 to 4 improves the Overall score by only 0.1 points, while increasing the maximum number of candidates for an interior token from 25 to 81. This suggests that reference-chain propagation already predicts positions close to useful matches, so further expanding the local window brings limited benefit.
Considering both accuracy and complexity, we set $r=2$ by default.

\section{Conclusion}
\label{sec:conclusion}

This paper presents a video codec-guided training-free token compression method for video large language models. The method exploits reference-frame dependencies, motion vectors, and residuals already available in compressed video to guide candidate construction during token compression. By aligning these coding signals with the visual token grid, the method reduces the search space for residual-guided token protection and motion-guided temporal merging without modifying the base video large language model. The key idea is to use codec information as a low-cost search prior. Residual information forms compact candidate pools for intra-frame token protection, and motion information constructs localized candidates for inter-frame token merging. This design reduces unnecessary feature comparisons while preserving the semantic reliability of visual token compression. Experiments on three video large language models and four video-understanding benchmarks show that the proposed method achieves performance comparable to, and in some cases better than, the state-of-the-art training-free token compression baseline FlashVID, while substantially reducing the feature comparisons required for token protection and temporal token merging. These results demonstrate that codec information already present in compressed video can effectively improve the accuracy-complexity tradeoff of video token compression.

\bibliographystyle{IEEEtran}
\bibliography{reference}
\begin{IEEEbiography}[{\includegraphics[width=1in,height=1.25in,clip,keepaspectratio]{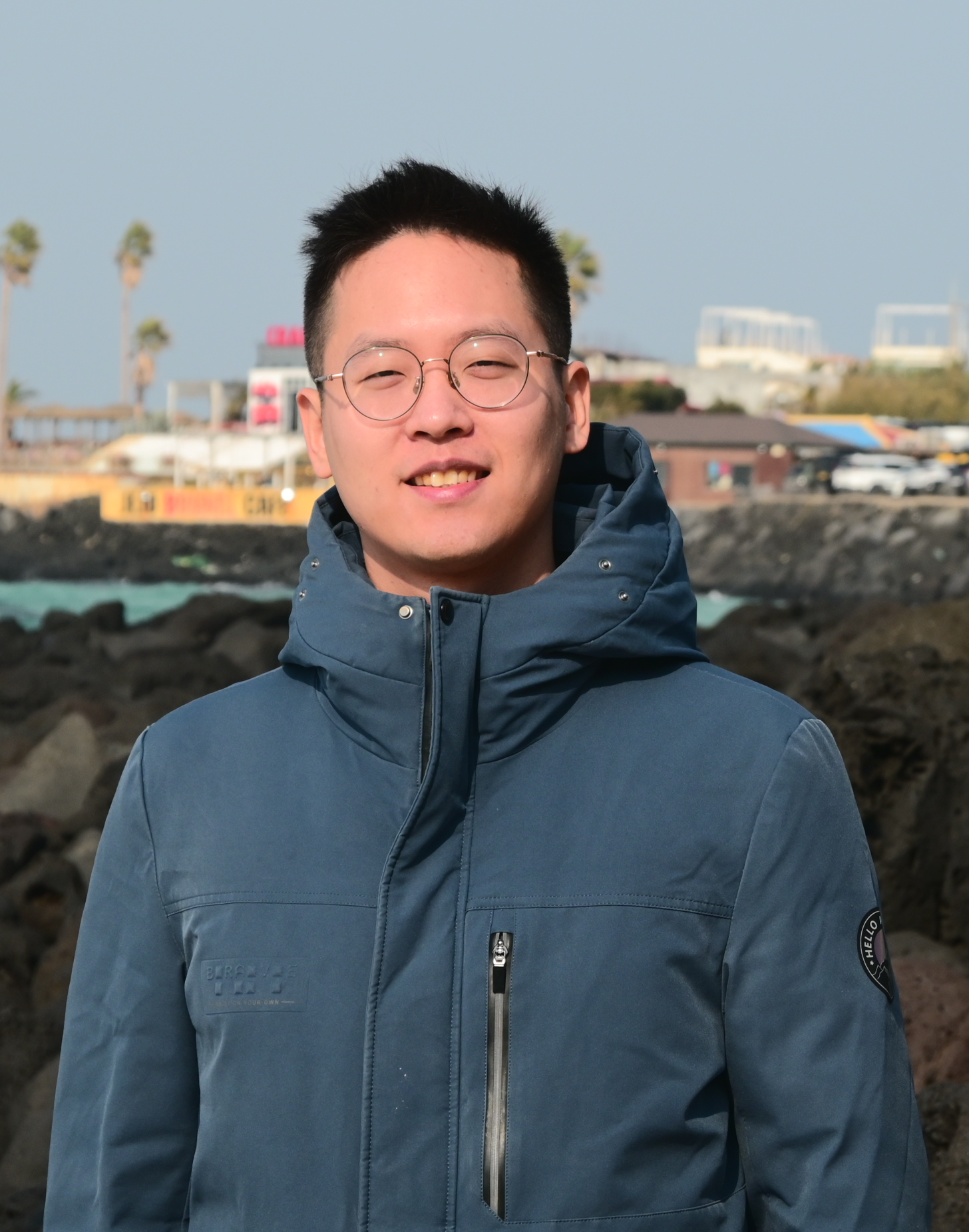}}]{Xihua Sheng} (Member, IEEE) received the B.S. degree in automation from Northeastern University, Shenyang, China, in 2019, and the Ph.D. degree in electronic engineering from University of Science and Technology of China (USTC), Hefei, Anhui, China, in 2024. 
He was a Postdoctoral Fellow at the department of computer science of City University of Hong Kong (CityU) from 2024 to 2026. He is currently a Research Assistant Professor  at the department of computing of The Hong Kong Polytechnic University (PolyU). 
His research interests include image/video/point cloud coding, signal processing, and machine learning.
\end{IEEEbiography}

\begin{IEEEbiography}[{\includegraphics[width=1in,height=1.25in,clip,keepaspectratio]{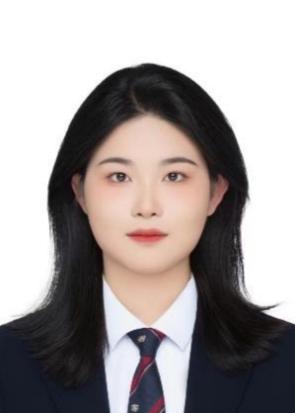}}]{Ruijie Liu} received the B.S. degree in Mathematics and Applied Mathematics from Jilin University, Changchun, China, in 2025. She is currently pursuing the M.Sc. degree with the Department of Informatics, Imaging and Data Sciences, The University of Manchester, Manchester, U.K.
\end{IEEEbiography}

\begin{IEEEbiography}[{\includegraphics[width=1in,height=1.25in,clip,keepaspectratio]{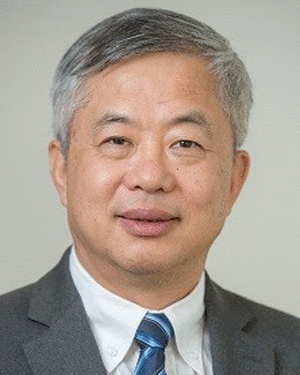}}]{Chang Wen Chen} (Fellow, IEEE) received the B.S. degree from the University of Science and Technology of China, in 1983, the M.S.E.E. degree from the University of Southern California, in 1986, and the Ph.D. degree from the University of Illinois at Urbana–Champaign, in 1992. He is currently chair professor of Visual Computing with The Hong Kong Polytechnic University. Before his current position, he served as dean of the School of Science and Engineering, The Chinese University of Hong Kong, Shenzhen from 2017 to 2020. He also served as an Empire Innovation professor with the University at Buffalo, the State University of New York from 2008 to 2021. He was Allen Henry Endow chair professor with the Florida Institute of Technology from 2003 to 2007. He was on the faculty with Electrical and Computer Engineering, University of Rochester from 1992 to 1996 and with the Faculty of Electrical and Computer Engineering, University of Missouri-Columbia from 1996 to 2003. He has served as the editor-in-chief for IEEE Trans. Multimedia from 2014 to 2016, and the editor-in-chief for IEEE Trans. Circuits and Systems for Video Technology from 2006 to 2009. He has been an editor for several other major IEEE Transactions and Journals, including as senior editor for the IEEE Journal of Selected Areas in Communications and the IEEE Journal of Selected Topics in Signal Processing. He has served as Conference chair for several major IEEE, ACM, and SPIE conferences related to multimedia communications and signal processing. His research has been funded by both government agencies and industrial corporations. He and his students have received 10 Best Paper Awards or Best Student Paper Awards more than the past two decades. He has also received several research and professional achievement awards. These include the Sigma Xi Excellence in Graduate Research Mentoring Award, in 2003, the Alexander von Humboldt Research Award, in 2009, the University at Buffalo Exceptional Scholar for Sustained Achievement Award, in 2012, the SUNY System Chancellor's Award for Excellence in Scholarship and Creative Activities, in 2016, and the University of Illinois ECE Distinguished Alumni Award, in 2019. He is an SPIE fellow (2007), and an elected member of the Academia Europaea (2021).
\end{IEEEbiography}
\end{document}